# Big Data and Education: using big data analytics in language learning

Vahid Ashrafimoghari
School of Business
Stevens Institute of Technology
Hoboken, NJ
vashraf1@stevens.edu

July 18, 2022

## Abstract

Working with big data using data mining tools is rapidly becoming a trend in education industry. The combination of the current capacity to collect, store, manage and process data in a timely manner, and data from online educational platforms represents an unprecedented opportunity for educational institutes, learners, educators, and researchers. In this position paper, we consider some basic concepts as well as most popular tools, methods and techniques regarding Educational Data Mining and Learning Analytics, and discuss big data applications in language learning, in particular.

## 1 Introduction

Big data and artificial intelligence applications in education have made significant headways in recent years. There is an increased interest and research in educational data mining, particularly in improving students' performance with various predictive and prescriptive models (Su et al., 2017; Khan & Alqahtani, 2020; Zhang et al., 2021). With the emergence of new technologies, new academic trends introduced into educational system which results in large data (Guleria & Sood, 2017). Today, students use the Internet to study and generate huge amounts of data that are not prepared by conventional learning methods. (Roy & Singh, 2017). Big Data technologies can process the large amount of data involved in education. Big Data Technology (BDT) plays an essential role in optimizing education intelligence by facilitating institutions, management, educators, and learners improved quality of education (Bamiah, Brohi & Rad, 2018). Big data technologies can improve the data acquisition, storage, processing, analysis, security, and virtualization for e-learning systems (Otoo-Arthur & van Zyl, 2020).

Educational Data Mining (EDM) and Learning Analytics (LA) are two growing fields of study to improve teaching and learning experience (Otoo-Arthur & van Zyl, 2019). LA and EDM provide a comprehensive background on adaptive learning. Big data can be used for the detection of attrition risk, data visualization, student's skill estimation, and grouping and collaborations among the students. Moreover, LA and EDM can identify learner interventions to make them more efficient and effective. Additionally, LA offers the possibility of implementing real-time assessment and feedback systems and processes at scale and can provide valuable insights into task design for instructors and materials designers (Chunzi, Xuanren & Ling, 2020).

In the following sections, we will briefly review the state of the art of EDM and LA. That will include discussion of the emerging fields of EDM and LA, tools, methods, and techniques commonly used in these fields, and the opportunities afforded by big data for language learning.

## 2  Educational Data Mining

Many experts believe that technology will not replace teachers, but rather will transform the role of teachers. In the past, teachers were the main source of information and knowledge. The future teacher will be more like a coach, helping students to find and use information for themselves. The educational technology industry is growing rapidly. Many schools and colleges are using educational technology to help students learn. Many new companies are developing new educational technology products. Many of these products are designed to help teachers teach more effectively. The future of educational technology is likely to be more and more data driven.

In general, Educational Data Mining (EDM) is a form of data mining that extracts knowledge from educational data. The most common types of data that are analyzed in EDM include student performance, student demographic, and student progress data. It can be used to discover patterns and trends in data, and to create models that can be used to predict future outcomes (Romero & Ventura, 2020).  As shown in Fig.1, the steps in this process are:
1. Collecting data from educational data sources
2. Cleaning and preprocessing the data
3. Mining the data for patterns and relationships
4. Generating new knowledge about education from the data

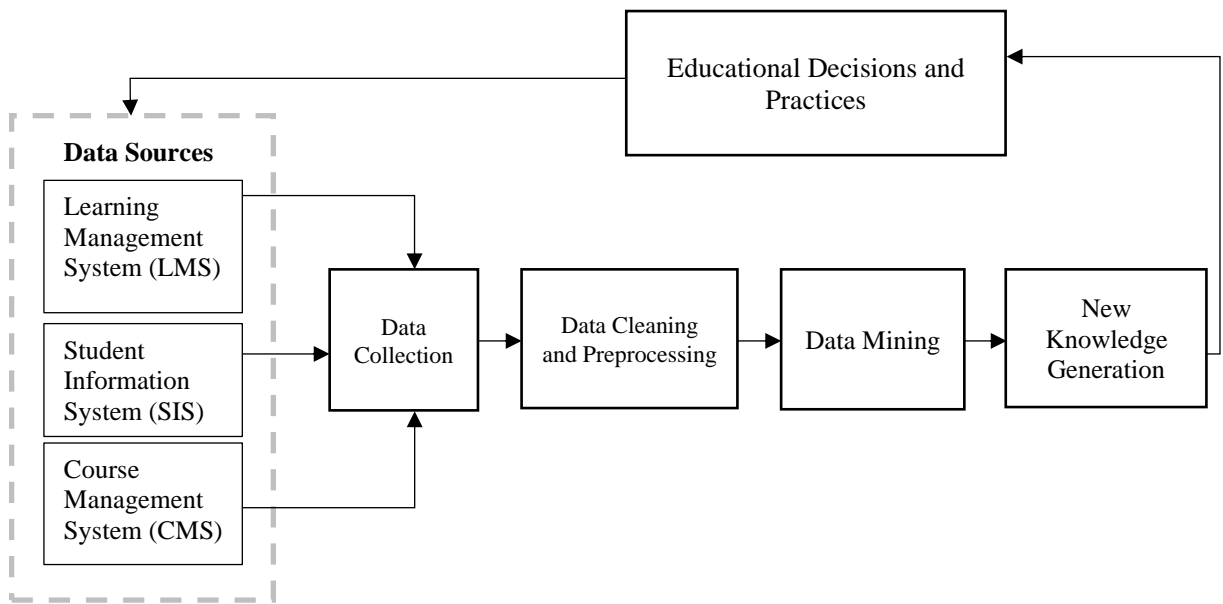

Fig.1- Educational Data Mining Process Diagram

Some of the examples of EDM are:

- Inferring students' mental models from their questions,
- Mining affective states from student interactions,
- Detecting plagiarism by comparing students' submitted work to online sources and other students' submissions,
- Building models to predict student performance on future exams, assignments, and other coursework,
- Identifying at-risk students based on patterns of behavior such as missed class sessions, late assignment submissions, etc.

# 3 Learning Analytics

Learning analytics (LA) can be defined as collecting, analyzing, representing, and reporting of data about learners and their contexts aiming for understanding and improving learning practices (Clow, 2013; Viberg et al., 2018). In other words, LA is a data-driven approach to understand how people learn, so that you can help them learn better. Here are a few examples of learning analytics:
- Tracking students' engagement with online course content,
- Predicting academic performance from social media data,
- Measuring the impact of different types of feedback on students' performance,
- Using data to inform decisions about which intervention strategies are most effective for struggling students.

There are different ways to approach LA, but one common framework comprises six main stages (as shown in Fig.2):

**1- Data Collection:** This includes all the data that can be collected such as student activity data, assessment data, and demographic data. This data can come from a variety of sources, including learning management systems (LMS), student information systems (SIS), surveys, interviews, focus groups, and observations A common approach is to use a data warehouse model, which collects data from multiple sources and then stores it in a central location where it can be easily accessed and analyzed. Another popular approach is to use a NoSQL database, which allows for more flexibility in terms of data collection and analysis.

**2- Data Analysis:** This refers to the process of analyzing the data to uncover patterns, trends, and insights. This can be done using a variety of methods, including descriptive statistics, predictive modeling, machine learning algorithms, text analytics, and social network analysis.

**3- Reporting/Data Visualization:** Once the data has been analyzed, it needs to be presented in a way that is easy to understand and use by decision-makers. This might involve creating reports, dashboards, or other visualizations that summarize the key findings.

**4- Decision Making:** The next step is to use the insights from the data analysis to make decisions about how to improve learning. This will involve setting goals for improvement, designing interventions, and selecting appropriate measures of success.

**5- Implementation:** The next step is to put the decisions into action by implementing the interventions and evaluating the results.

**6- Refinement:** The final step completes the cycle of LA by providing feedback to refine overall approach based on the results of implementation stages.

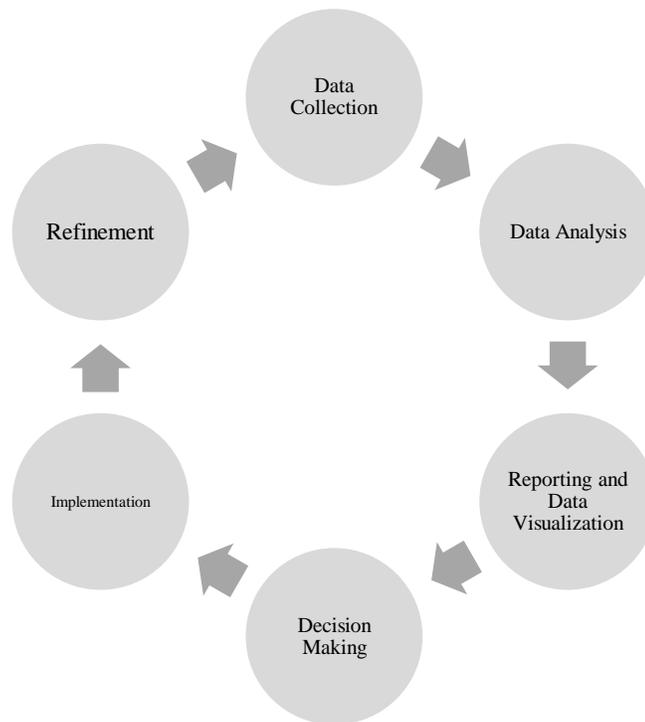

Fig.2- Learning Analytics Process Diagram

## 4 Tools, Methods, and Techniques

Some popular LA and EDM tools and platforms are listed in the Table.1. Each tool has its own strengths and weaknesses, so it's important to choose one that will best fit your needs. There is no one-size-fits-all solution for LA and EDM, as the best tool for a particular organization or individual may vary depending on factors such as budget, data availability, and desired features. These tools can help organizations track data such as learner engagement, completion rates, and test scores.

| Name | Description | Link |
|---|---|---|
| Knewton | It uses artificial intelligence and data science to personalize learning for students. Its commercial adaptive learning platform enables educational organizations to provide each learner with a personalized learning path that is constantly optimized based on the student's performance. | https://www.knewton.com/ |
| McGraw-Hill ConnectED | It is a commercial digital teaching and learning platform that gives students and teachers access to their course materials from any device. The platform includes all the | https://connected.mcgraw-hill.com/ |

| | resources needed for a course, including textbooks, assessments, and gradebook. | |
|---|---|---|
| Blackboard Learn | It is a commercial software application that allows educators to create and deliver course materials online. It also provides tools for collaboration, communication, and assessment. | https://www.blackboard.com/teaching-learning/learning-management/blackboard-learn |
| Datashop | It is a public repository of learning interaction data developed by CMU serving as a secure data repository for educational research. Recently, it has launched DataLab creating the world's largest bank of educational technology data. | https://pslcdatashop.web.cmu.edu/ |
| MeerkatED | It analyzes students' activity in a course offered on computer-supported collaborative learning tools. It prepares and visualizes overall snapshots of participants in the discussion forums, their interactions, and the leaders/peripheral students in the discussions and much more. | http://www.reirab.com/MeerkatED |
| Moodle Behaviour Analytics | It is a Moodle block plugin that is intended for extracting sequential behavior patterns of students from course access logs. | https://moodle.org/plugins/block_behaviour |
| Moodle Inspire | It is a Moodle plugin that implements open-source learning analytics using machine learning backends to provide predictions of learner success, and ultimately diagnosis and prescriptions (advisements) to learners and teachers. | https://moodle.org/plugins/tool_inspire |
| Yet Analytics | It provides commercial learning Record Store/ Data visualization tools. Also, it provides insightful analytics on talent development, role readiness and career pathing, rich engagement analytics through different learning ecosystems, and precise predictive analytics solutions. | https://www.yetanalytics.com/ |
| GISMO | It is a is a graphical interactive monitoring tool for Moodle that provides visualization of students' activities in online courses to instructors. | http://gismo.sourceforge.net/ |
| Wooclap | It is a commercial audience response system app that could be used to improve in-class students' engagement. It provides real-time analytics to professors and provides a feedback wall so that students can communicate (during the course and after) to their professors. | https://www.wooclap.com/ |
| Clever | It provides a commercial single sign-on tool to students and teachers in order to navigate between all software and learning resources and assists students in their learning process | https://clever.com/ |

| | and analyzes their personal engagement with learning resources. | |
|---|---|---|
| StREAM | It is a commercial student engagement analytics platform which provides educators with student engagement insight at cohort, course, module, and individual levels. | https://www.solutionpath.co.uk/stream/ |
| Bright Bytes | It provides a commercial SaaS-based data analytics solution that lets you evaluate how teachers and students use technology for learning, studies the availability of devices and Internet access throughout the school and at home, measures the skill levels of teachers and students with multimedia, and evaluates the school culture, professional development, and technology needs across the organization. | https://www.brightbytes.net/ |
| SmartKlass | It is a LA Moodle plug-in that should be included as a part of the Moodle virtual learning platform helping teachers to manage the learning journey of their students. | https://moodle.org/plugins/local_smart_klass |

Table 1- Most popular LA/EDM tools and platforms

Regardless of the differences between the LA and EDM areas, these two many common grounds in terms of the research objectives as well as methods and techniques that are used. Most methods applicable to educational data are employed in both EDM and LA. The most popular methods are predictive modeling, cluster analysis, and relationship mining. However, there are many other methods and techniques that cover a wide range of applications. The methods, their descriptions and a few examples are shown in Table 2.

| Name | Objective/Description | Application example |
|---|---|---|
| Predictive Modeling | It employs a number of methods and algorithms such as Neural Networks, Bayesian Networks, Logistic Regression and Support Vector Machines (SVM) to predict the future trends/events through identifying patterns of relationships in existing historical data. | Using Six data mining algorithms, a predictive model is developed to accurately predict fifth year and final Cumulative Grade Point Average (CGPA) of engineering students based on their program of study, the year of entry and the Grade Point Average (GPA) for the first three years of study (Adekitan and Salau, 2019). |
| Network Analytics | It is useful for identifying patterns of relationships or interactions between people, learning materials or other entities. | Using Social Network Analysis, each Problem-based Learning group was mapped, the level and structure of interaction among group participants |

|  |  | was determined, the isolated and the active students (leaders and facilitators) were identified, academic performance predicted, and the role of tutor was evaluated. (Saqr & Alamro, 2019) |
| --- | --- | --- |
| Visual LA | It refers to the graphical presentation of data and insights, either to aid in the analysis process or as a tool to communicate the results of analyses to various stakeholders. | The potential of using LA dashboards to improve the learning process and outcomes for both students and instructors were discussed (Han et al., 2021) |
| Text mining | This method uses computational linguistic techniques to generate insight into student writing, learning dialogues, teacher talk and other forms of educational text. | Using text mining techniques, a tool was developed to provide feedback to students on their use of rhetorical moves in their academic writing (Knight et al., 2020) |
| Causal mining | The objective of this method is to find causal relationships in data. (i.e., does repeatedly retrying quizzes cause decreased learning?) | Using causal mining the main drivers of success, or failure, of engineering students enrolled in an online course were investigated (de Carvalho et al., 2018) |
| Cluster analysis | It is a data analysis technique which employs various algorithms such as K-means to divide a dataset into groups of data (named clusters) within which data points have similar traits. | By classifying students into various categories based on their level of engagement, the researchers found that extrinsic factors are not enough to make students committed to Massive Open Online Courses (MOOCs), so adding intrinsic factors is recommended (Khalil & Ebner, 2017) |
| Process mining | The process mining employs data science to evaluate and improve workflows. | The University of Melbourne employed process mining to improve its admission workflow. As a result, its overall turnaround time reduced and its volume of applications increased, significantly (IEEE Task Force on Process Mining Case Study, 2020) |
| Relationship mining | It is a data mining technique that uses various approaches such as Graph Mining and Class Association Rules to look for patterns in relational databases | Using Class Association Rules combined with other data mining and statistical analysis approaches, the researchers showed that failure in an academic course presented in a blended learning environment, was associated with negative attitudes and perceptions of the students towards Moodle (Kotsiantis et al., 2013) |
| A/B testing | It is an experimental method to compare two versions of a single variable by testing a subject's response to variant A against variant B, finding which variant is more effective. | A/B testing and Randomized Control Trial (RCT) were used to evaluate the use of predictive data by teachers. The findings showed that teachers can positively impact students' |

| | | performance when predictive learning analytics is employed (Herodotou et al., 2019) |
|---|---|---|
| Knowledge domain modeling | This method incorporates knowledge of domain experts to add more value to raw data. | To improve existing e-tutoring systems, incorporating domain knowledge using ontology domain model to control adaptive e-tutor systems is recommended (Ahmed & Kovacs, 2020) |
| Discovery with models | It is a technique that uses a previously validated model of a phenomenon (i.e., through clustering) as a key component in another analysis. | The researchers developed a machine-learned detector of student carelessness and applied its outputs to motivational questionnaires, to discover the relationships between motivational measures and student carelessness (Harshkovitz et al., 2013) |

Table 2- Most popular LA/EDM methods and techniques

## 5 Big data and language learning

Some big data applications in language learning could include using data to track and study learner behavior, to better understand how people learn languages, to monitor general indicators such as attendance and performance on tests and specific language-related issues such as whether students achieve a certain number of target vocabulary items in a certain period, and to develop and improve language learning materials and methods (Hou, 2019). Additionally, big data could be used to monitor language learning progress in real-time and provide tailored feedback to learners, or to create personalized language learning journeys. This could include tracking the most frequent words and phrases by learners; observing the interaction of learners with language learning materials or monitor the progress of learners through different stages of their learning path. By tracking this data, educators could quickly identify areas where learners are struggling and provide them with targeted feedback or support. Moreover, this data could be used to improve language learning materials and methods by identifying areas that need more attention or focus. For instance, data on which languages are most spoken in a specific area can be used to tailor language learning materials to the needs of that population target. In another example, the data on how people use language learning tools can be used to improve the design of those tools. Data from online dictionaries and translation tools, for example, can be used to create more accurate and user-friendly tools (Dong et al, 2019; Chunzi et al., 2020; Zang & Wang, 2021).

Information about student learning can be obtained either during or outside of instructional time. The first is an example of "synchronous analytics" which can be used to track and analyze learning activities as they are happening (i.e., during a class session) in order to provide feedback and support in real-time. The learning activity can be either online or face-to-face. In online learning environment students need to be connected through a device. For example, a teacher might want to assess students' knowledge of the new vocabulary in the unit and compares the result with the other groups. "Asynchronous analytics" is the analysis of information outside of class session. This can be before or after an individual class, or before a course (to predict what will happen with a particular group of enrolled students compared with previous cohorts, based on specifics of those students such as their previous language knowledge, their backgrounds, their age, their attitude towards language learning, their motivation to learn a second language and so on) or after a course (to gain insight into the success of the course, for instance). An example would be to understand if there is a relationship between participation in group activities in the class and development of students' oral proficiency. Showing how these two are related could hugely help to increase engagement of inactive students and prompt them to communicate more often. Whereas the former is mostly used for the purposes of classroom management, to make decisions immediately affecting what happens in class, the latter is used for course and program management and usually has a longer time frame (Reinders & Lan, 2021; Wen & Song, 2021).

## 6 Challenges

There are a few challenges when it comes to using big data in education. Firstly, collecting accurate and complete data can be difficult and time-consuming. There are few databases available for educational data mining and learning analytics purposes and most of the platforms and software are commercial, so they are not publicly available which limits the ability of researchers to run big data projects in this domain. Secondly, analyzing big data can be complex and requires specialized skills. Since EDM and LA are interdisciplinary fields, the research team should comprise people with various skills from psychology to computer science to linguistics. Finally, ensuring data privacy and security is essential but can be difficult to achieve for educational data, in particular.

## 7 Conclusion

Educational institutions are applying EDM and LA to improve the services they provide by detecting areas for improvement, setting policies, and measuring results. With advances in adaptive learning systems, on one hand, it is possible to provide a personalized learning experience that fits learner's needs and capabilities. On the other hand, it helps educators to quickly see the effectiveness of their adaptations and interventions, providing feedback for continuous improvement. Researchers and developers can rapidly run A/B tests to compare versions A and B of material designs, educational products, and learning approaches, enabling the state of the art and the state of the practice to keep pace with the rapid pace of adoption of online and blended learning environments.

In this paper, the process and most common tools, methods and techniques used for EDM and LA briefly discussed. Where there is no one-size-fits-all solution for EDM and LA, the most popular methods are predictive modeling, cluster analysis, and relationship mining. Big data has a wide number of applications in language learning such as using data to track and study learner behavior at real-time, and to develop and personalize language learning materials and methods, and to improve equational systems and policies.

However, there are still some challenges to be addressed. Collecting accurate and complete data can be difficult and time-consuming, and still there is no data-driven culture in many educational institutes and comprehensive and user-friendly tools that could be integrated in the most popular LMSs are lacked.